\pdfoutput=1
\documentclass[11pt]{article}

\usepackage{acl}
\usepackage{enumitem}
\usepackage{times}
\usepackage{latexsym}
\usepackage{graphics}
\usepackage{multirow}
\usepackage[T1]{fontenc}
\usepackage[normalem]{ulem}
\usepackage{booktabs}
\usepackage{subcaption}
\usepackage{hyperref}
\usepackage{url}
\usepackage{graphicx}

\usepackage{soul}

\usepackage[utf8]{inputenc}

\usepackage{microtype}

\usepackage{float}
\usepackage{soul}
\usepackage{amsmath}

\definecolor{lightblue}{rgb}{.50,.90,0.51}
\definecolor{atomictangerine}{rgb}{1.0, 0.6, 0.4}

\newcommand{\xlmrob}{XLM-RoBERTa}
\newcommand{\bt}{BERT}
\newcommand{\subone}{\textit{subtask1}}
\newcommand{\subtwo}{\textit{subtask2}}
\newcommand{\subthree}{\textit{subtask3}}

\newcommand{\rashkincorpus}{\texttt{TSHP-17}}
\newcommand{\proppycorpus}{\texttt{QProp} }

\title{QCRI at SemEval-2023 Task 3: News Genre, Framing and Persuasion Techniques Detection using Multilingual Models}

\author{%
Maram Hasanain$^{1}$, Ahmed Oumar El-Shangiti$^{1}$\footnote{Work was done while on internship at QCRI}, Rabindra Nath Nandi$^2$, \\
\textbf{Preslav Nakov$^3$} and \textbf{Firoj Alam$^1$}
\\
$^1$Qatar Computing Research Institute, HBKU, Qatar\\
$^2$Hishab Singapore Pte. Ltd, Singapore, 
$^3$MBZUAI, UAE, \\
maramhasanain@gmail.com, ahmedmohamedlemin@gmail.com, \\rabindro.rath@gmail.com, preslav.nakov@mbzuai.ac.ae, fialam@hbku.edu.qa
}

\begin{document}
\maketitle
\begin{abstract}
% \mh{System paper submissions for a single task can be up to 5 pages. Acknowledgments, references, and appendices do NOT count toward page limits.}
Misinformation spreading in mainstream and social media has been misleading users in different ways. Manual detection and verification efforts by journalists and fact-checkers can no longer cope with the great scale and quick spread of misleading information. This motivated research and industry efforts to develop systems for analyzing and verifying news spreading online. The SemEval-2023 Task 3 is an attempt to address several subtasks under this overarching problem, targeting writing techniques used in news articles to affect readers' opinions. The task addressed three subtasks with six languages, in addition to three ``surprise'' test languages, resulting in 27 different test setups. This paper describes our participating system to this task. Our team is one of the 6 teams that successfully submitted runs for all setups. The official results show that our system is ranked among the top 3 systems for 10 out of the 27 setups.
%\hl{The average ranking of our runs across all 27 test subsets is 5.2}.
\end{abstract}

\section{Introduction}
\label{sec:introduction}
Monitoring and analyzing news have become an important process to understand how different topics (e.g., political) are reported in different news media and within and across countries. This has many important applications since the tone, framing, and factuality of news reporting can significantly affect public reactions toward social or political agendas. 
%, which often time can lead to chaos. In addition, monitoring live news media is important in different fields such as news on public health issues might be important as it may lead to a danger to the public, issues that are at the focus of public concerns, or news on global events (e.g., natural disasters, man-made disasters, political conflict)~\cite{steinberger2013multilingual}. 
%The information and/or news that is reported, and propagated has multifaceted aspects going beyond pure factuality. it can be an objective news with a certain framing dimension to change the focus of the audience and it is blended with propaganda techniques. An example of such news content is reported below in Figure \ref{}. Our study focused on addressing three different aspects: {\emph(i)} news genre categorization (i.e., news is objective, opinion or satire); {\emph(ii)} framing dimensions~\cite{card-etal-2015-media} and {\emph(iii)} propaganda techniques~\cite{da-san-martino-etal-2019-findings}. 
A news piece can be manipulated on multiple aspects to sway readers' perceptions and actions. Going beyond information factuality, other aspects include objectivity/genre, framing dimensions inserted to steer the focus of the audience~\cite{card-etal-2015-media}, and propaganda techniques used to persuade readers towards a certain agenda~\cite{BARRONCEDENO20191849,da-san-martino-etal-2019-findings}. %An example of such news content is shown in Figure~\ref{}. 

News categorization is a well studied problem in the natural language processing field. Recently, research attention has focused on classifying news by factuality~\cite{zhou2020survey,nakov2021survey}, or other related categorizations such as fake vs. satire news~\cite{LOW2022115824,golbeck2018fake}. However, there have been efforts towards other classification dimensions. \citet{card-etal-2015-media} developed a corpus of news articles annotated by 15 framing dimensions such as economy, capacity and resources, and fairness and equality, to support development of systems for news framing classification. Moreover, identifying propagandistic content has gained a lot of attention over several domains including news~\cite{BARRONCEDENO20191849,da-san-martino-etal-2019-findings}, social media~\cite{alam-etal-2022-overview} and multimodal content~\cite{dimitrov2021detecting,SemEval2021-6-Dimitrov}.

%The idea of media framing is to highlight, emphasize or obscure some aspects of the message over others in order to gain political, social or economic agendas. \citet{card-etal-2015-media} developed an annotated corpus of news articles 15 framing dimensions such as economy, capacity and resources, morality, fairness and equality, legality, constitutionality and jurisprudence, policy prescription and evaluation, crime and punishment, security and defense, health and safety, quality of life, cultural identity, public opinion, politics, external regulation and reputation, and other. This study shades the light in automatic detection of framing from news articles. The third aspect of our study is \textit{propaganda}. It is the expression of an opinion or an action by an individual or a group deliberately designed to influence the opinions or the actions of other individuals or groups with reference to predetermined ends, which is achieved by means of well-defined rhetorical and psychological devices. Propaganda techniques are commonly used in social media to manipulate or to mislead users. There has been efforts in identifying propogandistic content from news~\cite{BARRONCEDENO20191849,da-san-martino-etal-2019-findings}, social media~\cite{alam-etal-2022-overview} and multimodal content \cite{dimitrov2021detecting,SemEval2021-6-Dimitrov}.
% In recent years the propagandistic content has been able to spread in unprecedented manner thanks to the internet and the ease of posting on social media networks~\cite{BARRONCEDENO20191849,da-san-martino-etal-2019-findings}.

The SemEval-2023 Task 3 shared task aims at motivating research in the aforementioned categorization tasks, namely:  detection and classification of the \textit{genre}, \textit{framing}, and the \textit{persuasion techniques} in news articles~\cite{semeval2023task3}. It targets multiple languages including English, French, German, Italian, Polish, and Russian to push the research on multilingual systems. Moreover, to promote development of language-agnostic models, the task organizers released test subsets for three surprise languages (Georgian, Greek, and Spanish). 
%only at inference time, with no training data, which made it a more challenging problem.

%Our study focused on addressing three different aspects: {\emph(i)} news genre categorization (i.e., news is objective, opinion or satire); {\emph(ii)} framing dimensions~\cite{card-etal-2015-media} and {\emph(iii)} propaganda techniques~\cite{da-san-martino-etal-2019-findings}.

Our proposed system is based on fine-tuning transformer based models~\cite{vaswani2017attention} in multiclass and multi-label classification settings for different tasks and languages. 
% \mh{Need to write an overview of the system architecture here}. 
We participated in all three subtasks submitting runs for all nine languages, which resulted in 27 testing setups. We experimented with different mono and multilingual transformer models,
%In this paper, we present our systems ranked 2nd for  Detecting The Genre, The Framing, and The Persuasion technique in News Articles. Our models rely on state-of-the-art variations of transformer based models
 such as \bt{}~\cite{devlin2019bert} and~\xlmrob{}~\cite{roberta19,chi-etal-2022-xlm} among others. In addition, we also experimented with data augmentation. 

The rest of the paper is organized as follows. Section~\ref{sec:related_work} gives an overview of related work. In section~\ref{sec:system}, we present  the proposed system. In section~\ref{sec:experiments}, we provide the details of our experiments. Section~\ref{sec:results} presents the results for our official runs, and finally, we conclude our paper in section~\ref{sec:conclusion}.

\section{Related Work}
\label{sec:related_work}
%In this section, we review the related work on propaganda detection in NLP that has informed the approach taken in this research paper.

\subsection{News Genre Categorization}
% \mh{Ahmed}
Prior works on automated news categorization have focused on various aspects such as topic, style, how news is presented or structured, and intended audience~\cite{EINEA2019104076,chen2008web,yoshioka2001genre,stamatatos2000automatic}. %\citet{dai2018fine} defined four commonly used news article structures based on their selections and organizations of news elements. 
News articles have also been categorized based on their factuality and deceptive intentions~\cite{golbeck2018fake}. For example, fake news is false and the intention is deceive where satire news is also false but the intent is not deceive rather to call out, ridicule, or expose behavior that is shameful, corrupt, or otherwise ``bad''.

\subsection{Propaganda Detection}
%The detection of propaganda in natural language text has caught a significant attention from researchers in recent years~\cite{vijayaraghavan-vosoughi-2022-tweetspin}. 
%It 
Propaganda is defined as the use of automatic approaches to intentionally disseminate misleading information over social media platforms~\cite{woolley2018computational}. 
Recent work on propaganda detection has focused on news articles \cite{BARRONCEDENO20191849,rashkin-EtAl:2017:EMNLP2017,EMNLP19DaSanMartino,da2020survey}, multimodal content such as memes~\cite{dimitrov2021detecting,SemEval2021-6-Dimitrov} and tweets~\cite{vijayaraghavan-vosoughi-2022-tweetspin,alam-etal-2022-overview}. Several annotated datasets have been developed for the task such as \rashkincorpus{}~\cite{rashkin-EtAl:2017:EMNLP2017}, and \proppycorpus~\cite{BARRONCEDENO20191849}. \citet{Habernal.et.al.2017.EMNLP,Habernal2018b} developed a corpus with 1.3k arguments annotated with five fallacies (e.g., red herring fallacy), which directly relate to propaganda techniques. \citet{EMNLP19DaSanMartino} developed a more fine-grained taxonomy consisting of 18 propaganda techniques with annotation of news articles. Moreover, the authors proposed a multi-granular deep neural network that captures signals from the sentence-level task and helps to improve the fragment-level classifier. %Subsequently, a system was developed and made publicly available~\cite{da2020prta}. 
An extended version of the annotation scheme was proposed to capture information in multimodal content~\cite{dimitrov2021detecting}. Datasets in languages other than English have been proposed. For example, using the same annotation scheme from ~\cite{dimitrov2021detecting}, ~\citet{alam-etal-2022-overview} developed a dataset of Arabic tweets and organized a shared task on Arabic propaganda technique detection. \citet{vijayaraghavan-vosoughi-2022-tweetspin} developed a dataset of tweets, which are weakly labeled with different fine-grained propaganda techniques. They also proposed a neural approach for classification.

\subsection{Framing} 
Framing refers to representing different salient aspects and perspectives for the purpose of conveying the latent meaning about an issue~\cite{entman1993framing}. 
Recent work on automatically identifying media frames includes developing coding schemes and semi-automated methods~\cite{boydstun2013identifying}, 
datasets such as the Media Frames Corpus~\cite{card-etal-2015-media}, systems to automatically detect media frames \cite{liu2019detecting,zhang-etal-2019-tanbih}, large-scale automatic analysis of news articles~\cite{kwak2020systematic}, and semi-supervised approaches~\cite{cheeks2020discovering}. 

Given the multilingual nature of the datasets released with the task at hand, our work is focused on designing a multilingual approach for news classification for the three subtasks of interest.

\section{System Overview}
\label{sec:system}
Our system is comprised of preprocessing followed by fine-tuning pre-trained transformer models. The preprocessing part includes standard model specific tokenization. Our experimental setup consists of {\emph(i)} monolingual ($*_\text{mono}$): training and evaluating monolingual transformer model for each language and subtask; {\emph(ii)} multilingual ($*_\text{multi}$): combining subtask specific data from all languages for training, and evaluating the model on task and language specific data; {\emph(iii)} data augmentation ($*_\text{aug}$): applying data augmentation on language specific training set, then training a monolingual model using augmented dataset, and evaluating it on the test set. This has been applied for each subtask.  
% augmentation and multilingual. 
% Example names for tasks in the three setups: \subone$_\text{mono}$, \subone$_\text{multi}$, \subone$_\text{aug}$
% Suggested content: main system architecture and the 
% three setups we experimented with: monolingual, data augmentation and multilingual. Example names for tasks in the three setups: \subone$_\text{mono}$, \subone$_\text{multi}$, \subone$_\text{aug}$

\subsection{Data Augmentation}
\label{ssec:data_augmentation}
Data augmentation is an effective way to deal with class imbalance issues or to increase the size of the training dataset or increase within-class variation. 
%Linguistic data augmentation is naturally difficult than image augmentation considering the structural and semantic property of both data form. 
Typically, textual data augmentation has been done by upsampling techniques such as SMOTE~\cite{chawla2002smote}, however, that approach is applied to the vector representation. 
%It is challenging compared to the image augmentation due to the discrete nature of the language. 
Very recently, some useful strategies are introduced for textual data augmentation~\cite{feng2021survey}, which range from rule-based approaches to model-based techniques. \citet{wei2019eda} proposed a set of token-level random perturbation operations including random insertion, deletion, and swap, which have been employed in several studies~\cite{feng2021survey,alam2020punctuation}. 

We used such approaches with contextual representation from transformer models in this study. These include {\emph(i)} synonym augmentation using WordNet, {\emph(ii)} word insertion and substitution using BERT~\cite{devlin2019bert}, RoBERTa~\cite{liu2019roberta} and DistilBERT~\cite{sanh2019distilbert}. More details on the implementation of these approaches can be found in the following data augmentation package.\footnote{\url{https://github.com/makcedward/nlpaug}} 
% The idea was to 
% We believe these data augmentation techniques will be useful to our system to 
% help reduce the class imbalance in the training subsets and increase data size. 

% {\emph(iii)} word insertion using BERT, {\emph(iv)} word substitution using DISTILBERT and {\emph(v)} word substitution using ROBERTA. 
% to mitigate the class imbalance during training phase \cite{feng2021survey}. 
% We consider five types of augmentation techniques using  NLPAug data augmentation package\footnote{\url{https://github.com/makcedward/nlpaug}} during balancing the different class distribution: (i) synonym augmentation using WordNet, (ii) word substitution using BERT, (iii) word insertion using BERT, (iv) word substitution using DISTILBERT and (v) word substitution using ROBERTA.

\section{Experiments}
\label{sec:experiments}
%\mh{Maram}
In this section, we describe the tasks and datasets used during experiments and provide implementation details for our models.

\subsection{Task and Dataset}
The SemEval-2023 Task 3 is composed of 3 subtasks for each language: 
\begin{enumerate} %[itemsep=0pt, topsep=0pt, partopsep=0pt, parsep=0pt]
    \item \textbf{News Genre Categorization (\subone)}: Given a news article in a particular language, classify it to an \textit{opinion}, \textit{news reporting}, or a \textit{satire} piece. This is a multiclass classification task at the article level.
    
    \item \textbf{Framing Detection (\subtwo)}: Given a news article, identify the frames used in the article. This is a multi-label classification task at the article level. This task includes 14 frames/labels such as \textit{economic}, \textit{capacity and resources}, \textit{morality}, and \textit{fairness and equality}. %\textit{legality, constitutionality and jurisprudence}, \textit{policy prescription and evaluation}, \textit{crime and punishment}, \textit{security and defense}, \textit{health and safety}, \textit{quality of life}, \textit{cultural identity}, \textit{public opinion}, \textit{political}, \textit{external regulation and reputation} and \textit{other}.
    %This is a multi label classification task at the article level, where participants are asked to identify the frames used in an article. 14 Frames were considered.
    \item \textbf{Persuasion Techniques Detection (\subthree)}: Given an article, identify the persuasion technique(s) present in each paragraph. This is a multi-label classification task at the paragraph level. This task includes 23 techniques/labels such as \textit{loaded language}, \textit{appeal to authority}, \textit{appeal to popularity}, and \textit{appeal to values.}%,, appeal to fear/prejudice, flag waving, causal oversimplification, false dilemma or no choice, consequential oversimplification, straw man, red herring, whataboutism, slogans, appeal to time, conversation killer, repetition, exaggeration or minimisation, obfuscation - vagueness or confusion, name calling or labeling, doubt, guilt by association, appeal to hypocrisy, and questioning the reputation. 
\end{enumerate}

\begin{table}
  \centering
  \resizebox{\columnwidth}{!}{%
 \begin{tabular}{l|c}
    \hline
{\bf HF Model Name} & {\bf Language} \\
\hline
xlm-roberta-large & Multilingual\\
bert-large-cased & English \\
\textbf{roberta-large} & English\\
dbmdz/bert-base-french-europeana-cased & French\\
dbmdz/bert-base-german-uncased & German \\
\textbf{uklfr/gottbert-base} & German\\
dbmdz/bert-base-italian-uncased & Italian\\
sdadas/polish-roberta-large-v2 & Polish\\
\textbf{allegro/herbert-large-cased} & Polish\\
DeepPavlov/rubert-base-cased & Russian \\
\hline
\end{tabular}%
}
\caption{Pre-trained models used in experiments. For languages with multiple models, the best ones are shown in bold, which are also comparable in the monolingual training setup on the dev subset across all three subtasks.}
\label{tab:models}
\end{table}

The task organizers released three subsets (train, development and test) of data per language of the six main languages for each subtask. Further details and statistics can be found in~\cite{semeval2023task3}. Starting with the six \textit{train} subsets, we apply three methods to acquire new versions of these train subsets:
\begin{enumerate}
    \item Train subset splitting: we randomly split each of the train subsets into 80-20 splits to acquire training and validation subsets for each subtask and each language. As will be shown in the following subsection, our models were re-trained using different random seeds. The validation set is used to select the random seed leading to the best model.
    \item Multilingual dataset construction: to support our multilingual training setup, we combine the training subsets resulting from the previous step for all languages to create a multilingual training subset. We apply the same approach to the validation subsets.
    \item Data augmentation: %\hl{
    for each of our generated training  splits, we apply data augmentation to it and use the resulting datasets to train a monolingual model for each subtask and each language.
    %}
\end{enumerate}

\begin{table}[h]
\resizebox{\columnwidth}{!}{%
\begin{tabular}{c|r|l|cc}
\hline
\textbf{Lang}  & \textbf{Rank} & \textbf{Run} & \textbf{F1$_\text{macro}$} & \textbf{F1$_\text{micro}$} \\
\hline
\multirow{3}{*}{EN} & 1 & MELODI & 0.784 & 0.815 \\
 & 16 & Baseline & 0.288 & 0.611 \\
 & \textbf{17} & QCRI$_\text{multi}$ & 0.281 & 0.593 \\ \hline
\multirow{3}{*}{FR} & 1 & UMUTeam & 0.835 & 0.880 \\
 & \textbf{2} & QCRI$_\text{aug}$ & 0.767 & 0.800 \\
 & 10 & Baseline & 0.568 & 0.740 \\ \hline
\multirow{4}{*}{GE} & 1 & UMUTeam & 0.820 & 0.820 \\
& 1 & SheffieldVeraAI & 0.820 & 0.820 \\
 & \textbf{7} & QCRI$_\text{mono}$ & 0.667 & 0.660 \\
 & 9 & Baseline & 0.630 & 0.760 \\ \hline
 
\multirow{3}{*}{IT} & 1 & Hitachi & 0.768 & 0.852 \\
 & \textbf{7} & QCRI$_\text{mono}$ & 0.541 & 0.787 \\
 & 12 & Baseline & 0.389 & 0.672 \\ \hline
\multirow{3}{*}{PO} & 1 & FTD & 0.786 & 0.936 \\
 & \textbf{10} & QCRI$_\text{mono}$ & 0.571 & 0.830 \\
 & 13 & Baseline & 0.490 & 0.830 \\ \hline
\multirow{3}{*}{RU} & 1 & Hitachi & 0.755 & 0.750 \\
 & \textbf{6} & QCRI$_\text{multi}$ & 0.567 & 0.653 \\
 & 12 & Baseline & 0.398 & 0.653 \\ \hline
\multirow{3}{*}{KA} & 1 & Riga & 1.000 & 1.000 \\
 & \textbf{4} & QCRI$_\text{multi}$ & 0.622 & 0.897 \\
 & 13 & Baseline & 0.256 & 0.345 \\ \hline
\multirow{3}{*}{GR} & 1 & SinaaAI & 0.806 & 0.813 \\
 & \textbf{4} & QCRI$_\text{multi}$ & 0.708 & 0.813 \\
 & 15 & Baseline & 0.171 & 0.344 \\ \hline
\multirow{3}{*}{ES} & 1 & DSHacker & 0.563 & 0.567 \\
 & \textbf{3} & QCRI$_\text{multi}$ & 0.489 & 0.567 \\
 & 16 & Baseline & 0.154 & 0.300  \\         
\hline
\end{tabular}%
}
  \caption{Official results for all nine test languages in \subone. \textbf{F1$_\text{macro}$} is the official evaluation measure for this subtask. Subscripts for our team runs indicate the training setup used.}
  \label{sub1:results}
\end{table}

\subsection{Implementation Details}
We use HuggingFace (HF) library~\citep{wolf-etal-2020-transformers} on top of PyTorch framework
 \cite{paszke2017automatic} as our base and source of all the pre-trained language models. Since different random initialization can considerably affect the model performance, we train the model for each language with $k$ different random seeds. 

For all experiments, we use Adam optimizer~\cite{adam_2015} with the learning rate of 2×10$^{-5}$. In setting other parameters of the models, we distinguish between \subone{} and \subtwo{} that operate on the document level, and \subthree$_\text{multi/aug}$ that works at the paragraph level and has a much larger training subset. Only for \subthree$_\text{multi/aug}$, the number of epochs=5, $k$=5, maximum sequence length=256, and batch size=8. For all remaining training setups and subtasks, the number of epochs=10, $k$=10, maximum sequence length=512, and batch size=4.  

For each of the three training setups described in section~\ref{sec:system}, the models trained using $k$ seeds for a language are evaluated over our validation subset using the official evaluation measure for the corresponding subtask. The model with the best performance is then applied to the development set. Eventually, the training setup that has the best performance on the development subset will be used to generate the official run for the corresponding subtask and test language. As for the ``surprise'' test languages, we use the model trained on the multilingual training subset with the best performance on the multilingual validation subset. 

For our multilingual training setup, we opt to use XLM-RoBERTa~\cite{roberta19}. As for all other setups, we used per-language monolingual pre-trained models listed in Table~\ref{tab:models}.

\section{Results}
\label{sec:results}
The results for our official runs per subtask are shown in Tables~\ref{sub1:results},~\ref{sub2:results} and~\ref{sub3:results}. For each subtask, we compare our official runs to two baselines: the top run in each test language, and the baseline as reported by the task organizers.

\begin{table}[]
\resizebox{\columnwidth}{!}{%
\begin{tabular}{c|r|l|cc}
\hline
\textbf{Lang}  & \textbf{Rank} & \textbf{Run} & \textbf{F1$_\text{micro}$} & \textbf{F1$_\text{macro}$} \\
\hline
\multirow{3}{*}{EN} & 1 & SheffieldVeraAI & 0.579 & 0.539 \\
 & \textbf{7} & QCRI$_\text{multi}$ & 0.513 & 0.419 \\
 & 18 & Baseline & 0.350 & 0.274 \\ \hline
\multirow{3}{*}{FR} & 1 & MarsEclipse & 0.553 & 0.537 \\
 & \textbf{7} & QCRI$_\text{multi}$ & 0.480 & 0.430 \\
 & 15 & Baseline & 0.329 & 0.276 \\ \hline 
\multirow{3}{*}{GE} & 1 & MarsEclipse & 0.711 & 0.660 \\
 & \textbf{2} & QCRI$_\text{multi}$ & 0.660 & 0.606 \\
 & 17 & Baseline & 0.487 & 0.418 \\ \hline
\multirow{3}{*}{IT} & 1 & MarsEclipse & 0.617 & 0.545 \\
 & \textbf{2} & QCRI$_\text{multi}$ & 0.599 & 0.479 \\
 & 13 & Baseline & 0.486 & 0.372 \\ \hline
\multirow{3}{*}{PO} & 1 & MarsEclipse & 0.673 & 0.638 \\
 & \textbf{3} & QCRI$_\text{multi}$ & 0.642 & 0.599 \\
 & 10 & Baseline & 0.594 & 0.532 \\ \hline
\multirow{3}{*}{RU} & 1 & MarsEclipse & 0.450 & 0.303 \\
 & \textbf{3} & QCRI$_\text{multi}$ & 0.434 & 0.364 \\
 & 13 & Baseline & 0.230 & 0.218 \\ \hline
\multirow{3}{*}{KA} & 1 & SheffieldVeraAI & 0.654 & 0.679 \\
 & \textbf{6} & QCRI$_\text{multi}$ & 0.517 & 0.457 \\
 & 13 & Baseline & 0.260 & 0.251 \\ \hline
\multirow{3}{*}{GR} & 1 & SheffieldVeraAI & 0.546 & 0.454 \\
 & \textbf{6} & QCRI$_\text{multi}$ & 0.519 & 0.400 \\
 & 13 & Baseline & 0.345 & 0.057 \\ \hline
\multirow{3}{*}{ES} & 1 & mCPT & 0.571 & 0.455 \\
 & \textbf{6} & QCRI$_\text{multi}$ & 0.488 & 0.390 \\
 & 17 & Baseline & 0.120 & 0.095 \\
\hline
\end{tabular}%
}
\caption{Official results for all nine test languages in \subtwo. \textbf{F1$_\text{micro}$} is the official evaluation measure for this subtask. Subscripts for our team runs indicate the training setup used.}
\label{sub2:results}
\end{table}

\begin{table}[h]
\resizebox{\columnwidth}{!}{%
\begin{tabular}{c|r|l|cc}
\hline
\textbf{Lang}  & \textbf{Rank} & \textbf{Run} & \textbf{F1$_\text{micro}$} & \textbf{F1$_\text{macro}$} \\
\hline
\multirow{3}{*}{EN} & 1 & APatt & 0.376 & 0.129 \\
 & \textbf{8} & QCRI$_\text{multi}$ & 0.320 & 0.133 \\
 & 19 & Baseline & 0.195 & 0.069 \\ \hline
\multirow{3}{*}{FR} & 1 & NAP & 0.469 & 0.322 \\
 & \textbf{5} & QCRI$_\text{multi}$ & 0.401 & 0.226 \\
 & 16 & Baseline & 0.240 & 0.099 \\ \hline
\multirow{3}{*}{GE} & 1 & KInITVeraAI & 0.513 & 0.233 \\
 & \textbf{3} & QCRI$_\text{multi}$ & 0.498 & 0.231 \\
 & 17 & Baseline & 0.317 & 0.083 \\ \hline
\multirow{3}{*}{IT} & 1 & KInITVeraAI & 0.550 & 0.214 \\
 & \textbf{6} & QCRI$_\text{multi}$ & 0.513 & 0.209 \\
 & 16 & Baseline & 0.397 & 0.122 \\ \hline
\multirow{3}{*}{PO} & 1 & KInITVeraAI & 0.430 & 0.179 \\
 & \textbf{5} & QCRI$_\text{multi}$ & 0.378 & 0.156 \\
 & 18 & Baseline & 0.179 & 0.059 \\ \hline
\multirow{3}{*}{RU} & 1 & KInITVeraAI & 0.387 & 0.189 \\
 & \textbf{3} & QCRI$_\text{multi}$ & 0.361 & 0.182 \\
 & 15 & Baseline & 0.207 & 0.086 \\ \hline
\multirow{3}{*}{KA} & 1 & KInITVeraAI & 0.457 & 0.328 \\
 & \textbf{2} & QCRI$_\text{multi}$ & 0.414 & 0.339 \\
 & 14 & Baseline & 0.138 & 0.141 \\ \hline
\multirow{3}{*}{GR} & 1 & KInITVeraAI & 0.267 & 0.126 \\
 & \textbf{2} & QCRI$_\text{multi}$ & 0.265 & 0.129 \\
 & 14 & Baseline & 0.088 & 0.006 \\ \hline
\multirow{3}{*}{ES} & 1 & TeamAmpa & 0.381 & 0.244 \\
 & \textbf{4} & QCRI$_\text{multi}$ & 0.350 & 0.157 \\
 & 11 & Baseline & 0.248 & 0.020 \\    
\hline
\end{tabular}%
}
  \caption{Official results for all nine test languages in \subthree. \textbf{F1$_\text{micro}$} is the official evaluation measure for this subtask. Subscripts for our team runs indicate the training setup used.}
  \label{sub3:results}
\end{table}

We observe that the multilingual models are generally the best performing models across all tasks. On average, the performance of the system was best for \subthree{} with a slight average ranking difference compared to \subtwo. Another interesting observation is that although \subthree{} has much larger train subsets , since it operates on the paragraph level, this did not improve the average system ranking across languages when compared to \subtwo. The results also clearly show the robustness of our model across languages and subtasks, as it managed to be among the best 3 runs for 10 out of the 27 test subsets, and it was among the top 5 runs for 15 of them.

Results over \subone{} and \subthree{} showed that our proposed system had a strong cross-lingual transfer ability when training the model on multilingual data and testing it on unseen languages (Georgian, Greek and Spanish).

\section{Conclusion}
\label{sec:conclusion}
In this paper, we presented our experiments and findings on news genre categorization, framing and persuasion techniques detection on multiple languages, which was a part of SemEval-2023 Task 3 shared task. The task includes 27 test setups for three subtasks and nine test languages. Our team successfully submitted runs for all setups. We proposed a system that is based on fine-tuning transformer models in multiclass and multi-label classification settings. We experimented with different mono and multilingual pre-trained models, in addition to data augmentation. From the experimental results, we observed that our multilingual model based on \xlmrob{} performs better across all tasks, even on unseen languages. 

Our future work includes domain adaptation and further exploration of data augmentation techniques. 

\section*{Ethics and Broader Impact}

\paragraph{Biases}
We note that there might be some biases in the data we use, however, we used the data that organizers made available. The biases, in turn, will likely be exacerbated by the unsupervised models trained on them. This is beyond our control, as the potential biases in pre-trained large-scale transformers models, which we use in our experiments.

\section*{Acknowledgments}
This publication was made possible by NPRP grant 14C-0916-210015 \emph{The Future of Digital Citizenship in Qatar: a Socio-Technical Approach} from the Qatar National Research Fund. 

Part of this work was also funded by Qatar Foundation's IDKT Fund TDF 03-1209-210013: \emph{Tanbih: Get to Know What You Are Reading}.

% The findings herein are solely the responsibility of the authors.
The views, opinions, and findings presented in this paper are those of the authors alone and do not necessarily reflect the views, policies, or positions of the QNRF or any other affiliated organizations.

\bibliography{bib/anthology,bib/propaganda,bib/bibliography}

\appendix

%\section{Appendix}
%\label{sec:appendix}
%This is an appendix.

\end{document}